\documentclass[letterpaper, 10 pt, conference]{ieeeconf}  

\IEEEoverridecommandlockouts                              

\usepackage{graphics} 
\usepackage{epsfig} 
\usepackage{mathptmx} 
\usepackage{times} 
\usepackage{amsmath} 
\usepackage{amssymb}  
\usepackage{fancyvrb}
\usepackage{commath}
\usepackage{xcolor}
\usepackage[nodots]{numcompress}
\usepackage{fleqn,multirow,graphicx,wrapfig,lineno,hyperref}
\hypersetup{colorlinks=true,linkcolor=blue}
\graphicspath{ {figs/} }

\title{\LARGE \bf
Robot-Assisted Drilling on Curved Surfaces with Haptic Guidance under Adaptive Admittance Control*
}

\author{Alireza Madani$^{1}$, Pouya P. Niaz$^{1}$, Berk Guler$^{1}$, Yusuf Aydin$^{2}$, and Cagatay Basdogan$^{1}$
\thanks{*This work was supported by the Scientific and Technological Research Council of Turkey (TUBITAK) under contract number EEEAG-117E645}
\thanks{$^{1}$A.M., P.P.N., B.G., and C.B. \textit{(Corresponding Author)} are with the Robotics and Mechatronics Laboratory and KUIS AI-Center,
        Koc University, Istanbul, Turkey, 34450
        {\tt\small {\{amadani20, pniaz20, berkguler20, cbasdogan\}@ku.edu.tr}}}%
\thanks{$^{2}$Y.A. is with the Department of Electrical and Electronics Engineering, MEF University,
        Istanbul, Turkey, 34396
        {\tt\small aydiny@mef.edu.tr}}%
}

\begin{document}

\maketitle
\thispagestyle{empty}
\pagestyle{empty}

\begin{abstract}

%
Drilling a hole on a curved surface with a desired angle is prone to failure when done manually, due to the difficulties in drill alignment and also inherent instabilities of the task, potentially causing injury and fatigue to the workers. On the other hand, it can be impractical to fully automate such a task in real manufacturing environments because the parts arriving at an assembly line can have various complex shapes where drill point locations are not easily accessible, making automated path planning difficult. In this work, an adaptive admittance controller with 6 degrees of freedom is developed and deployed on a KUKA LBR iiwa 7 cobot such that the operator is able to manipulate a drill mounted on the robot with one hand comfortably and open holes on a curved surface with haptic guidance of the cobot and visual guidance provided through an AR interface. Real-time adaptation of the admittance damping provides more transparency when driving the robot in free space while ensuring stability during drilling. After the user brings the drill sufficiently close to the drill target and roughly aligns to the desired drilling angle, the haptic guidance module fine tunes the alignment first and then constrains the user movement to the drilling axis only, after which the operator simply pushes the drill into the workpiece with minimal effort. Two sets of experiments were conducted to investigate the potential benefits of the haptic guidance module quantitatively (Experiment I) and also the practical value of the proposed pHRI system for real manufacturing settings based on the subjective opinion of the participants (Experiment II). 
The results of Experiment I, conducted with 3 naive participants, show that the haptic guidance improves task completion time by 26\% while decreasing human effort by 16\% and muscle activation levels by 27\% compared to no haptic guidance condition. The results of Experiment II, conducted with 3 experienced industrial workers, show that the proposed system is perceived to be easy to use, safe, and helpful in carrying out the drilling task. 
\\

\end{abstract}


\section{INTRODUCTION}

In the near future, humans and robots are expected to perform collaborative tasks involving physical interaction in various different environments such as homes, hospitals, and factories.

To this end, impedance and admittance controllers are often used for regulating the physical interaction between the human and the robot (pHRI). In admittance control, the input is the interaction force between human, robot, and the environment, and the output is the reference velocity which the robot is to follow \cite{Duchaine2007,duchaine2009safe}. 



One of the application areas where pHRI can be used for improving task efficiency is the manufacturing industry. Robots have been used in industrial settings at an increasing rate since the 1980s as a result of the improvements in their capabilities and ease of use \cite{Heyer2010,Ben-Ari2017}.
Many large-scale manufacturing tasks have therefore been successfully automated thanks to these advances in robotics. \textcolor{black}{In the meantime,} several small-batch tasks requiring higher level decision-making and supervision are still performed by humans. In recent years though, pHRI has enabled humans to collaborate with robots, resulting in such tasks being done more efficiently \cite{Ajoudani2017,Selvaggio2021}. 

\begin{figure}[t]
	\includegraphics[width=\columnwidth]{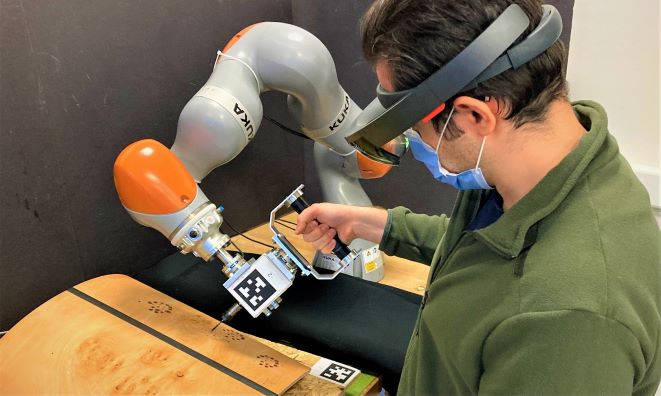}
	\centering
	\caption{Our pHRI system designed for drilling on a curved surface consists of a cobot, a drill attached to it, two force sensors, and a Hololens AR goggle.}
	\label{fig:6Dsetup}
	  \vspace{-1.5em}
\end{figure}

In fact, the number of pHRI studies that focus on improving task performance, practicality, completion time and ergonomics in manufacturing industries has increased dramatically during the last few years \cite{Matheson2019,Hu2020}. For instance, Cherubini et al. \cite{Cherubini2016a} used pHRI for mitigating muscular pain in an automotive assembly line. Zhang et al. \cite{Zhang2022} proposed a framework for task planning in pHRI settings for minimizing fatigue. Lamon et al. \cite{Lamon2020} introduced a visuo-haptic guidance interface for mobile manipulators to understand human instructions better in manufacturing environments. Guerin et al. \cite{Guerin2014} proposed a pHRI framework with different levels of autonomy for collaborative tasks in manufacturing settings. Later on, they used a similar framework to make pHRI more convenient in tasks such as bending metallic pieces and spot welding, based on predefined sets of robot capabilities \cite{Guerin2015}. Ochoa et al. \cite{Ochoa2021} proposed an impedance control scheme that captured human skills in a glass mold microdrilling task, then programmed a robot to automate the drilling. Kana et al. \cite{Kana2021} also proposed an impedance control scheme for pHRI-enabled curve tracing techniques in edge chamfering and polishing. Perez-Ubeda et al. \cite{Perezubeda2019} used the milling of soft materials as a practical example for determining the capabilities and limitations of collaborative robots in machining applications. 

Despite many studies on automated robotic drilling in the literature \cite{Frommknecht2017,Bu2017}, the number of studies on robotic-assisted drilling with physical human guidance is limited. However, fully automated drilling does not always obviate the demands of numerous realistic scenarios in manufacturing settings such as opening a hole on a curved piece in which the robot does not \textcolor{black}{have any prior knowledge of what may be located beneath the drilling surface with which collision should be avoided}. In another scenario, we have the human trying to keep a piece \textcolor{black}{of another assembly part} on the workpiece and drill both of them together. Under all such circumstances, the human operator \textcolor{black}{will not be able to} work alongside a pre-programmed robot. In the meantime, manual drilling, requiring the stabilization of the workpiece with fixtures, is time-consuming yet not precise enough. Accounting for the abovementioned rationales, fusing the human intelligence and supervision with robot's robustness and precision is the key to our concerns.
%
Aydin et al. \cite{Aydin2020} worked on a collaborative drilling task \textcolor{black}{by identifying the suitable control parameters of an admittance controller and adapting them} for a safe \textcolor{black}{and stable} operation. Sirintuna et al. \cite{Sirintuna2020} utilized a fractional-order admittance control scheme to better balance the trade-off between transparency (minimal resistance to the human) during free motion and stability during drilling with a cobot. Both of the drilling studies mentioned above only considered drilling on a flat surface using a single degree of freedom admittance controller. 

In this study, a pHRI system was developed for performing small-batch manufacturing tasks with haptic guidance of cobot, visual feedback provided through an AR interface, and an interaction controller for admittance adaptation. 
Drilling holes on a curved surface with a desired angle was selected as a case study to demonstrate the potential use of the proposed pHRI system in manufacturing settings (Fig. \ref{fig:6Dsetup}). It is difficult to open a hole on a curved surface at a custom angle using a manual drill. On the other hand, even with full knowledge of drill point location and drilling angle, workpieces with different curvature and material properties may arrive at the station in an assembly line, some with protrusions and cavities, some with sensitive or loose pieces attached to them, making full automation unfeasible. Hence, we propose to use a cobot for haptic guidance, weight compensation, and precise alignment, an augmented reality (AR) interface for visual guidance, and an adaptive controller for admittance adaptation to assist the human operator in opening holes on a curved surface. We therefore aim to fuse the intellectual capacity of the human (where, at what angle, and how deep the hole should be drilled, as well as how the robot should be guided to that location) with the mechanical precision and robustness of the robot to execute the task. 
In our drilling scenario, the user brings the drill attached to the robot sufficiently close to the target point on the curved surface utilizing an augmented reality (AR) interface, and then the robot aligns the drill bit with the selected drilling angle using the so-called \emph{haptic guidance} module, allowing the human to drill the selected hole with exact desired angle by simply pushing the drill forwards to the target point.

\textcolor{black}{In this work,} two sets of human experiments were conducted \textcolor{black}{aiming} to evaluate the effectiveness of haptic guidance in task performance, and to \textcolor{black}{assess} the user's \textcolor{black}{willingness to accept the technology}, as well as feasibility of the proposed system. The first experiment was conducted with three naive participants who had no prior experience with our system. Their drilling task performance under haptic guidance condition was compared with that of no haptic guidance condition using some quantitative metrics including accuracy in drill orientation, task completion time, and effort made by the operator. In the second experiment, the focus was on the subjective experience of the user about the proposed system under haptic guidance condition. This experiment was performed with three industrial workers who had extensive experience in drilling. Following the experiments, the participants were asked to fill out a questionnaire regarding their subjective opinions about the system and its potential use in real manufacturing environments.

In this study, an adaptive 6-DoF admittance controller is used for drilling a curved surface. The adaptation mechanism enables a good balance between transparency (minimal resistance to the human) and stability (see \cite{Aydin2018a, Aydin2020a, Sirintuna2020}). It keeps the pHRI system transparent as long as the robot is driven freely in space. Once it is sufficiently close to the target point (i.e. within a predefined radius), the robot becomes stiffer to prevent collision with the curved workpiece and help the human with the rough alignment of the drill bit. For the final drilling phase, after the robot aligns the drill bit precisely and confines the human movement to the desired drilling axis, the robot becomes even stiffer, to maximize stability during the drilling to open a hole.

The remainder of this paper is organized as follows:

\hyperref[sec:approach]{Section 2} presents our approach including the hardware components, the control architecture, and the details of our drilling application. \hyperref[sec:exp1]{Section 3} presents our first experiment, where we compare the task performance of 3 naive participants when they perform the drilling task with and without haptic guidance. \hyperref[sec:exp2]{Section 4} presents our second experiment, where we examine the feasibility of deploying the proposed system in real manufacturing settings based on the subjective opinion of 3 well-experienced industrial workers. Finally, \hyperref[sec:conclusion]{Section 5} \textcolor{black}{discusses} the experimental results and presents our \textcolor{black}{final remarks}.

\section{APPROACH}
\label{sec:approach}

We propose a solution for the task of drilling holes on curved surfaces with potentially unknown geometries, at locations marked by the user and for user-defined drilling angles. To this end, we developed a pHRI system whose components and their inner workings are explained below in detail.

\subsection{Hardware}

\begin{figure}[h]
	\includegraphics[width=\columnwidth]{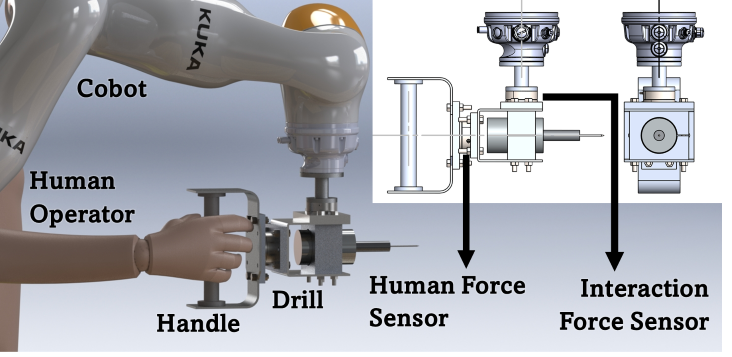}
	\centering
	\caption{A closer shot of the hardware components used in our study.}
	\label{fig:drillsetup}
\end{figure}

The major hardware components of our system are (1) a cobot: a 7R, KUKA LBR iiwa 7 R800 robot, (2) a powered drill: includes a DC motor operable between 0 and 48 Volts, and a drill bit, (3) two ATI Mini45 force/torque sensors: one of them is attached between the drill and the robot's end-effector, measuring interaction forces applied to it, while the other is attached between the drill and the handle which is held by the operator, measuring human force, and (4) an AR goggle: a Microsoft Hololens augmented reality interface (Fig. \ref{fig:drillsetup}).

\subsection{Control Loop}

\begin{figure}[h]
	\includegraphics[width=\columnwidth]{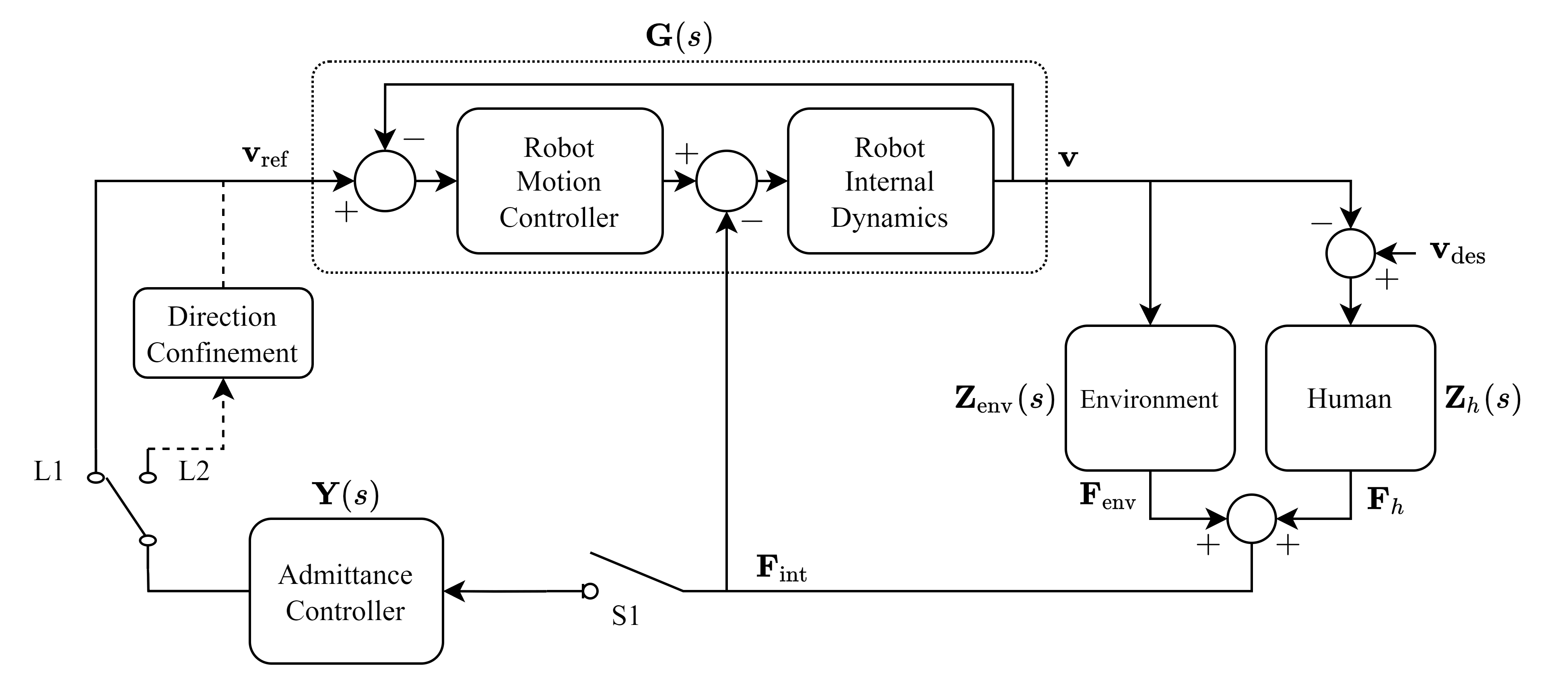}
	\centering
	\caption{Closed-loop control system used in our pHRI study.}
	\label{fig:6Dcontrol}
\end{figure}

The closed-loop control system used in this study is shown in Fig. \ref{fig:6Dcontrol}. 
Here, $\mathbf{G}(s)$ is the transfer function of the robot and its internal controller, whose output is the actual velocity of the robot's end-effector $\mathbf{v}$. \textcolor{black}{The} mechanical impedance of human and environment are shown as $\mathbf{Z}_h(s)$ and $\mathbf{Z}_\mathrm{env}(s)$ respectively. Resultant of \textcolor{black}{the} environmental force $\mathbf{F}_\mathrm{env}$ and the force applied by human $\mathbf{F}_h$ is the interaction force $\mathbf{F}_\mathrm{int}$, which is sent to the admittance controller $\mathbf{Y}(s)$, whose output is the reference end-effector velocity $\mathbf{v}_\mathrm{ref}$. 
%
In Fig. \ref{fig:6Dcontrol}, $\mathbf{v}_\mathrm{des}$ is the desired velocity of the human operator. When there is no haptic guidance, the operator has to manually align the drill with the desired drill orientation. Therefore, switches S1 and L1 are closed. However, when haptic guidance is activated, switch S1 and L1 are closed only when driving the robot in free space. When the robot reaches the 5-cm vicinity of the drill target, it locks for the automatic alignment, at which point switch S1 is opened. \textcolor{black}{During a 4 second interval during which the robot aligns itself} with the desired drilling orientation, the admittance control is not active. After alignment, switches S1 and L2 are closed, admittance control is reactivated, and the robot's motion is confined to the drilling axis, so that the operator can push the drill forwards to \textcolor{black}{the} drill target to open a hole.

In Fig. \ref{fig:6Dcontrol}, all variables are 6D vectors, including three translational terms and three rotational terms. For example, the velocity vectors in the figure are expressed as $\mathbf{v} = [v_x \: v_y \: v_z \: \omega_x \: \omega_y \: \omega_z]^\top$ where $v$ and $\omega$ stand for linear and angular velocities, respectively. The same applies to \textcolor{black}{wrench} vectors in the figure, which include 3 components of force and 3 components of torque, with respect to the three orthogonal axes.

\subsection{Admittance Controller}

We utilize 6 decoupled admittance controllers, one for each degree of freedom, to regulate the translational and rotational interactions between the robot and the human. For each translational degree of freedom, the admittance controller can be expressed in the Laplace domain as below:

\begin{equation}
    Y(s) = \frac{V_{\mathrm{ref}}(s)}{F_{\mathrm{int}}(s)} = \frac{1}{ms+b}
    \label{eq:admittancecontroltranslation}
\end{equation}

\noindent where $V_{\mathrm{{ref}}}(s)$ is the reference velocity in the Laplace domain that is given by the controller, and will be followed by the robot, $F_{\mathrm{{int}}}(s)$ is the input interaction force in the Laplace domain, $m$ and $b$ are the admittance mass and damping respectively, and $s$ is the Laplace variable. A similar relationship can be written between torque and reference angular velocity for each rotational degree of freedom as well. 
%


\subsection{Extracting Surface Information}

The scenario explored in this study includes a curved workpiece with an arbitrary shape arriving at the station, after which the operator marks a point on it for drilling, and selects an appropriate drilling angle (not necessarily perpendicular). In order to open a hole at the marked point (drill target) with a desired drilling axis, the location of the drill target and the tangent plane at the drill target must be known with respect to the base frame of the robot. For this purpose, a 3D surface mesh model of the workpiece was obtained by a depth camera first and then registered with the real surface using a virtual tag attached to it (Fig. \ref{fig:6Dsetup}).
Alternatively, the drill target and a few points around it can be sampled with the tip of the drill bit manually to register the target location and to estimate the tangent plane at the target location. Though time consuming and not as accurate as optical scanning techniques, this approach, which was tested by us, eliminates the need for a complete 3D surface mesh model of the curved surface. 

\subsection{Haptic Guidance}

\begin{figure}[h]
	\includegraphics[width=0.3\columnwidth]{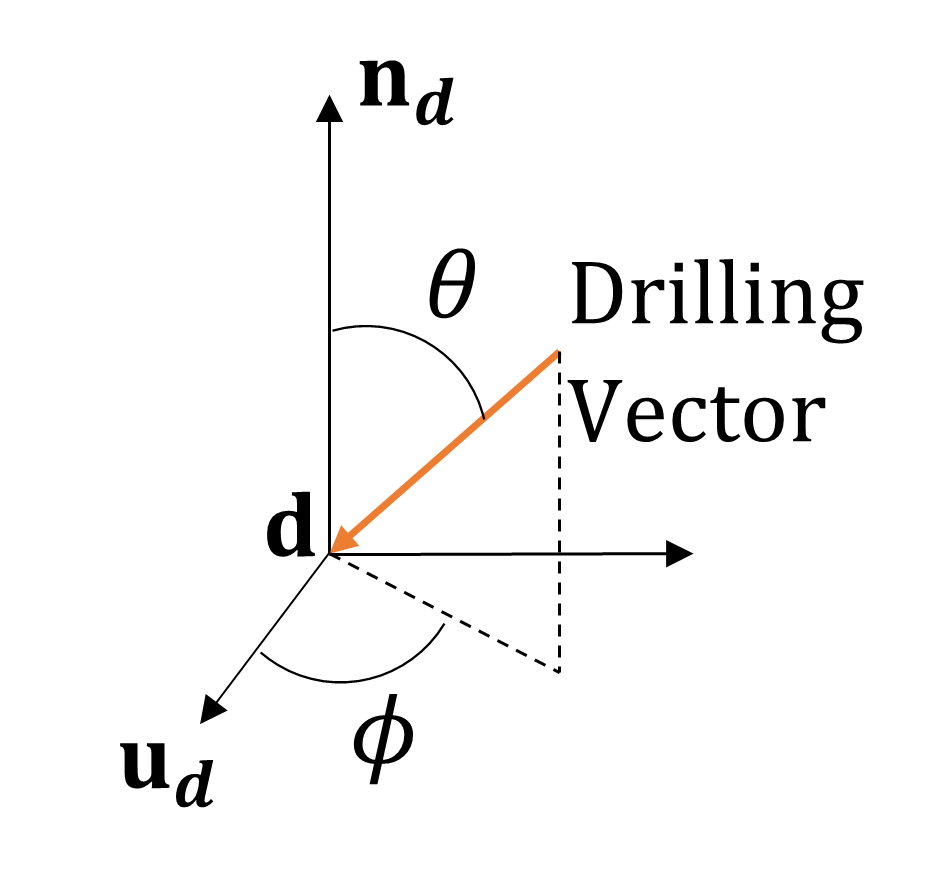}
	\centering
	\caption{Polar ($\phi$) and azimuth ($\theta$) angles are required for defining the drilling axis. To this end, the surface normal $\mathbf{n}_d$ and another auxiliary vector $\mathbf{u}_d$ normal to it are required for constructing a coordinate system at the drill target $\mathbf{d}$.}
	\label{fig:sphericalcoords}
	\vspace*{-\baselineskip}
\end{figure}

Using the tangent plane at the drill target, the surface normal vector can be estimated and a coordinate frame, whose origin is at the drill target, with one of its axes \textcolor{black}{being} the surface normal and another axis is parallel to the tangent plane at the target point, can be defined. 
After defining a coordinate frame at the target point, any arbitrary drilling axis can be \textcolor{black}{defined} with respect to it. As shown in Fig. \ref{fig:sphericalcoords}, \textcolor{black}{a} vector for \textcolor{black}{the} drilling axis can be expressed \textcolor{black}{by} a \textcolor{black}{pair of} polar ($\phi$) and azimuth ($\theta$) angles. When the operator selects the desired angle and brings the drill sufficiently close to the drill target and roughly aligns the drill bit using the visual feedback provided by the AR interface, the robot can align the drill bit precisely and quickly with the selected angle. After this alignment, the user movement (and hence the movement of the drill bit) is restricted to the drilling vector (axis) only. The user can then simply push the drill bit forward to open a hole on the curved surface in a safe manner. 

\subsection{Admittance Adaptation}

In order to cope with the varying requirements of the task in its different phases, adaptive admittance control is utilized in this study.
As long as the drill bit is far from the drill target, low admittance damping is used to keep the system transparent to human input, so the robot can be driven faster in free space. To prevent accidental contacts with the workpiece and help the user roughly align the drill bit in a more controlled manner, medium admittance damping was used when drill tip is sufficiently close (10 cm) to the target. Once the drill bit is at the 5-cm vicinity of the target, the robot takes over the control and aligns the drilling axis precisely. After this point, the user is constrained to move along the drilling axis only under high admittance damping for maximum stability during drilling to open a hole. \textcolor{black}{It is noteworthy that transition of damping values follows a ramp profile with a length of 1 second. Using the analysis conducted in our earlier work \cite{GULER2022102851} we observed that this will not lead to instability of the pHRI system.}  The values used for admittance damping in this study are tabulated in Table~\ref{tab:6Ddampvals}.

\begin{table}[h]
\centering
\caption{Control parameters used in the 6D drilling experiments.}
\resizebox{\columnwidth}{!}{%
\begin{tabular}{|l|l|l|l|l|l|}
\hline
\textbf{Phase} &
  \textbf{\begin{tabular}[c]{@{}l@{}}Degrees of \\ Freedom\end{tabular}} &
  \textbf{\begin{tabular}[c]{@{}l@{}}Translational\\ Mass\\ {[}kg{]}\end{tabular}} &
  \textbf{\begin{tabular}[c]{@{}l@{}}Translational\\ Damping\\ {[}Ns/m{]}\end{tabular}} &
  \textbf{\begin{tabular}[c]{@{}l@{}}Rotational\\ Mass\\ {[}kgm\textsuperscript{2}{]}\end{tabular}} &
  \textbf{\begin{tabular}[c]{@{}l@{}}Rotational\\ Damping\\ {[}Nms{]}\end{tabular}} \\ \hline
Free Motion        & 6 & 50 & 100  & 10 & 5  \\ \hline
Close to Target & 6 & 50 & 600  & 10 & 20 \\ \hline
Drilling          & 1 & 50 & 1000 & -- & -- \\ \hline
\end{tabular}%
}
\label{tab:6Ddampvals}
\vspace*{-\baselineskip}
\end{table}

\subsection{Augmented Reality Interface}

\begin{figure}[h]
	\includegraphics[width=\columnwidth]{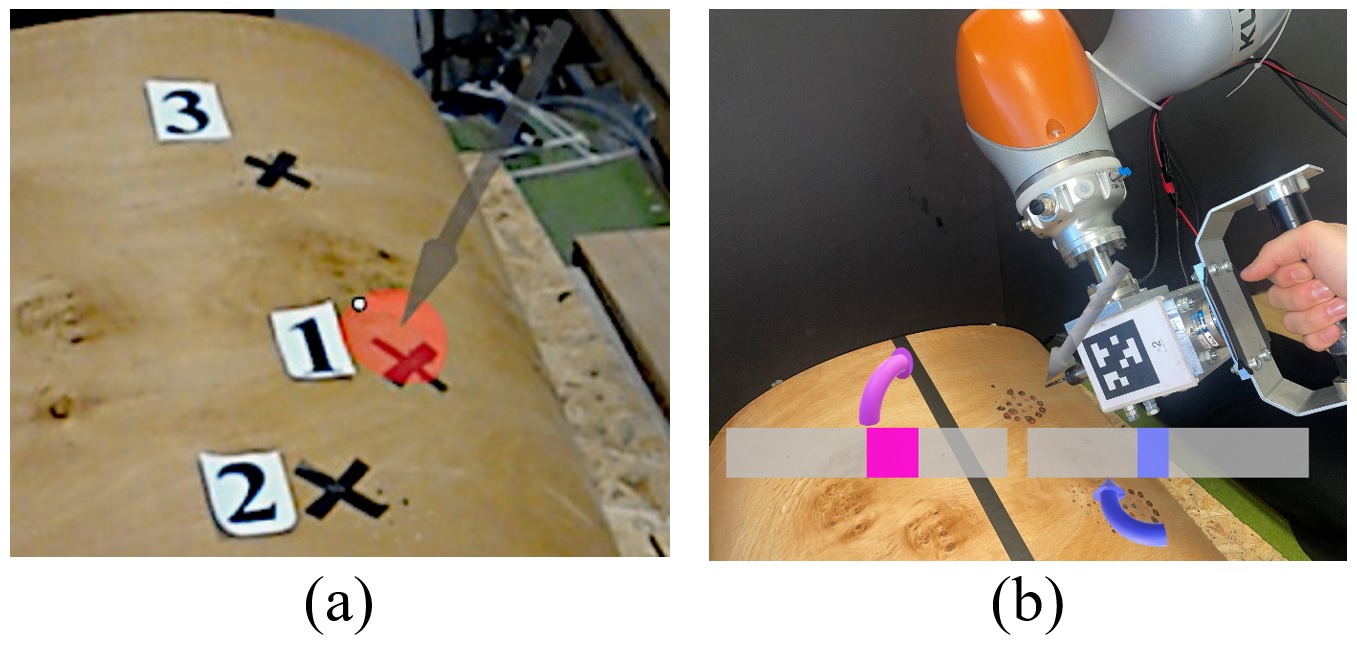}
	\centering
	\caption{The visual feedback displayed to the user by the augmented reality (AR) interface under (a) haptic guidance and (b) no haptic guidance}
	\label{fig:gogglevisual}
\end{figure}

The AR interface used in this study assists the operator during the drilling task by superimposing visual images on top of the real ones (Fig. \ref{fig:gogglevisual}). In a manufacturing environment involving a scenario such as ours, it can be difficult for the operator to remember the steps of the whole task and the sequence of drill targets. In our implementation, the text and images displayed through the AR interface guides the user throughout the task.  Furthermore, it can be hard to visualize how one should roughly align the drill bit in 3D space before the final and precise alignment made by the robot. A white colored and semi-transparent arrow with its tip pointing to the drill target is displayed through the AR interface to help the user with the rough alignment and reduce the task execution time. Furthermore, the AR interface highlights the current drill target by displaying a red colored hemisphere at its location (Fig. \ref{fig:gogglevisual}a).
This hemisphere also informs the operator about how close the drill tip \textcolor{black}{is} to the target before the robot takes over the control for the finer alignment of the drilling axis.




\section{EXPERIMENT I}
\label{sec:exp1}

In this experiment, the goal was to investigate the potential benefits of haptic guidance in task performance by comparing it with no guidance condition (i.e. manually aligning the drill bit with the desired drilling angle). Three naive subjects with an average age of $29.7 \pm 4.6$ participated in this experiment. \textcolor{black}{To monitor the} muscle activation levels of participants during the task execution, surface electromyography (sEMG) sensors were attached to their Flexor Carpi muscle. The procedures for sensor placement and data acquisition are identical to those presented in \cite{Sirintuna2020a}.

\subsection{Experimental Conditions}

There were two experimental conditions in this experiment: 1) haptic guidance by the robot and limited visual guidance by the AR interface and 2) no haptic guidance by the robot but more visual guidance by the AR interface with respect to the first condition. Under the second condition, the participant had to manipulate the drill manually to bring it to the drill target and align it with the desired drilling angle. Because there was no automated alignment, the system was not restricted to 1D motion in the drilling phase, unlike in the haptic guidance condition (condition 1). \textcolor{black}{Under this condition, again for giving haptic feedback to the operator and easing the subtle orientation adjustments, as soon as the drill bit gets closer than 10 cm to the drilling point, the damping increases to its high value (b = 1000 Ns/m). This ensures safe and stable contact interaction with our workpiece as well.} Also, there was no red hemisphere displayed on top of the drill target in condition 2 since the participants needed to visually see the exact target location without any visual obstacles on the way. The white transparent arrow remained still active under this condition, pointing to the drill target and showing the axis with which the participants should align the robot as precisely as they can. 
Furthermore, to compensate for the lack of haptic guidance in condition 2, the orientation errors with respect to the desired azimuth and polar angles were displayed as horizontal bars, which were updated in real-time as the user rotated the drill bit. Finally, two virtual arrows (one for the azimuth angle and one for polar angle) showed which way the drill needs to be rotated to get closer to the desired orientation. The values for the parameters used in this experiment are tabulated in Table \ref{tab:6DexpDetails}.

\subsection{Protocol}

\begin{figure}[h]
	\includegraphics[width=\columnwidth]{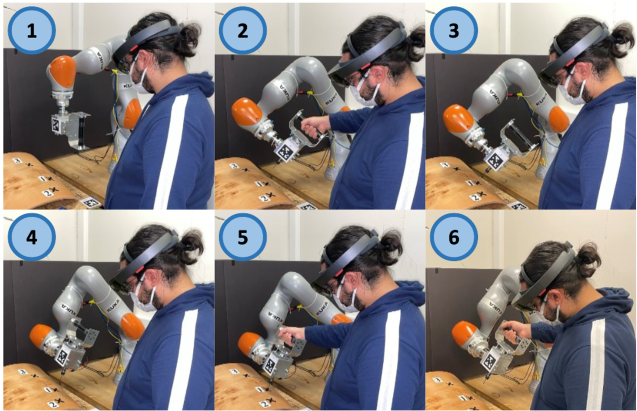}
	\centering
	\caption{Steps followed by the participants in Experiment I under haptic guidance condition.
	}
	\label{fig:deneyprotokolu}
\end{figure}

We first present the protocol followed by the participants under the haptic guidance condition below (see Fig. \ref{fig:deneyprotokolu}).\footnote{A demonstrative video of the proposed system can be found  at \url{https://www.yusuf-aydin.com/iros-2022-robot-assisted-drilling/}} 


\begin{enumerate}
    \item The participant approaches the robot, and stands on a marked space on the ground, waiting for the system to \textcolor{black}{initialize} (Fig. \ref{fig:deneyprotokolu}:1).
    \item \textcolor{black}{After being told to do so}, the participant grabs the handle of the robot, and moves the drill bit close to a red transparent hemisphere shown in the AR goggle (Fig. \ref{fig:gogglevisual}), while roughly aligning it with a white transparent arrow (i.e. desired drilling axis) displayed through the AR interface (Fig. \ref{fig:deneyprotokolu}:2).
    \item When the drill tip touches the red hemisphere, the robot takes over the control, while the message ``Please do not touch the robot" is displayed to the participant through the visor of the AR interface. The participant stands clear of the robot (Fig. \ref{fig:deneyprotokolu}:3).
    \item The robot aligns the drill bit with the drilling axis automatically in 4 seconds (Fig. \ref{fig:deneyprotokolu}:4).
    \item When the alignment is over and the robot is ready to be guided towards the target for drilling, the user is restricted to 1D forwards/backwards movements along the drilling axis. At this point, the message ``Continue..." is displayed to the participant.
    \item The participant grabs the handle again and simply \textcolor{black}{pushes} the drill into the workpiece to open a hole. The haptic guidance provided by the robot keeps the movement on the drilling axis. After drilling the hole, the participant retracts the drill out of the workpiece by moving back along the same axis again (Fig. \ref{fig:deneyprotokolu}:5-6).
\end{enumerate}

\textcolor{black}{After the first point is drilled, the user is ready to go to the next drilling point.} The 6-DoF admittance controller is reactivated, the hemisphere and arrow are displayed at the next target location, and the same steps are repeated for the new drill target.

Under condition 2 (no haptic guidance), the participants were asked to align the drill manually as accurately as possible using the white transparent arrow pointing to the drill target only. Moreover, the red hemisphere was not displayed to the participants through the AR interface at the drill target to prevent the obstruction of their view. \textcolor{black}{Furthermore, to assist the user during alignment, angular deviations from the desired drilling angles were displayed through the AR interface in the form of real-time error bars as well as guiding arrows which help them correct those deviations} (Fig. \ref{fig:gogglevisual}b). These features (horizontal bars and guiding arrows) were not displayed to the participants under haptic guidance condition (Fig. \ref{fig:gogglevisual}a). 


%
%

Before the actual experiments, participants were given two training trials, executed with and without haptic guidance, to familiarize themselves with the setup and the experimental procedures, without opening a hole on the workpiece. Then, each participant performed two trials of each condition, amounting to 4 trials in total; two trials with haptic guidance and two without haptic guidance.

\vspace*{-\baselineskip}
\begin{table}[h]
\centering
\caption{Parameters used in Experiment I.}
\resizebox{\columnwidth}{!}{%
\begin{tabular}{|l|l|l|}
\hline
\textbf{Item}                                                               & \textbf{Value} & \textbf{Unit} \\ \hline
Control adaptation distance from drill target                         & 10             & cm             \\ \hline
Locking distance from the drill target (radius of red hemisphere) & 5              & cm             \\ \hline
Distance from target after alignment                                   & 5              & cm             \\ \hline
Autopilot alignment time                                                   & 4              & sec         \\ \hline
First drill target polar angle $\phi_1$                                 & 5              & deg         \\ \hline
First drill target azimuth angle $\theta_1$                                  & 0              & deg         \\ \hline
Second drill target polar angle $\phi_2$                              & 30             & deg         \\ \hline
Third drill target polar angle $\phi_3$                              & 45             & deg         \\ \hline
Second and third drill target azimuth angles $\theta_2$ and $\theta_3$    & 10             & deg         \\ \hline
\end{tabular}%
}
\label{tab:6DexpDetails}
\end{table}
\vspace*{-\baselineskip}
\begin{figure}[b]
	\includegraphics[width=\columnwidth]{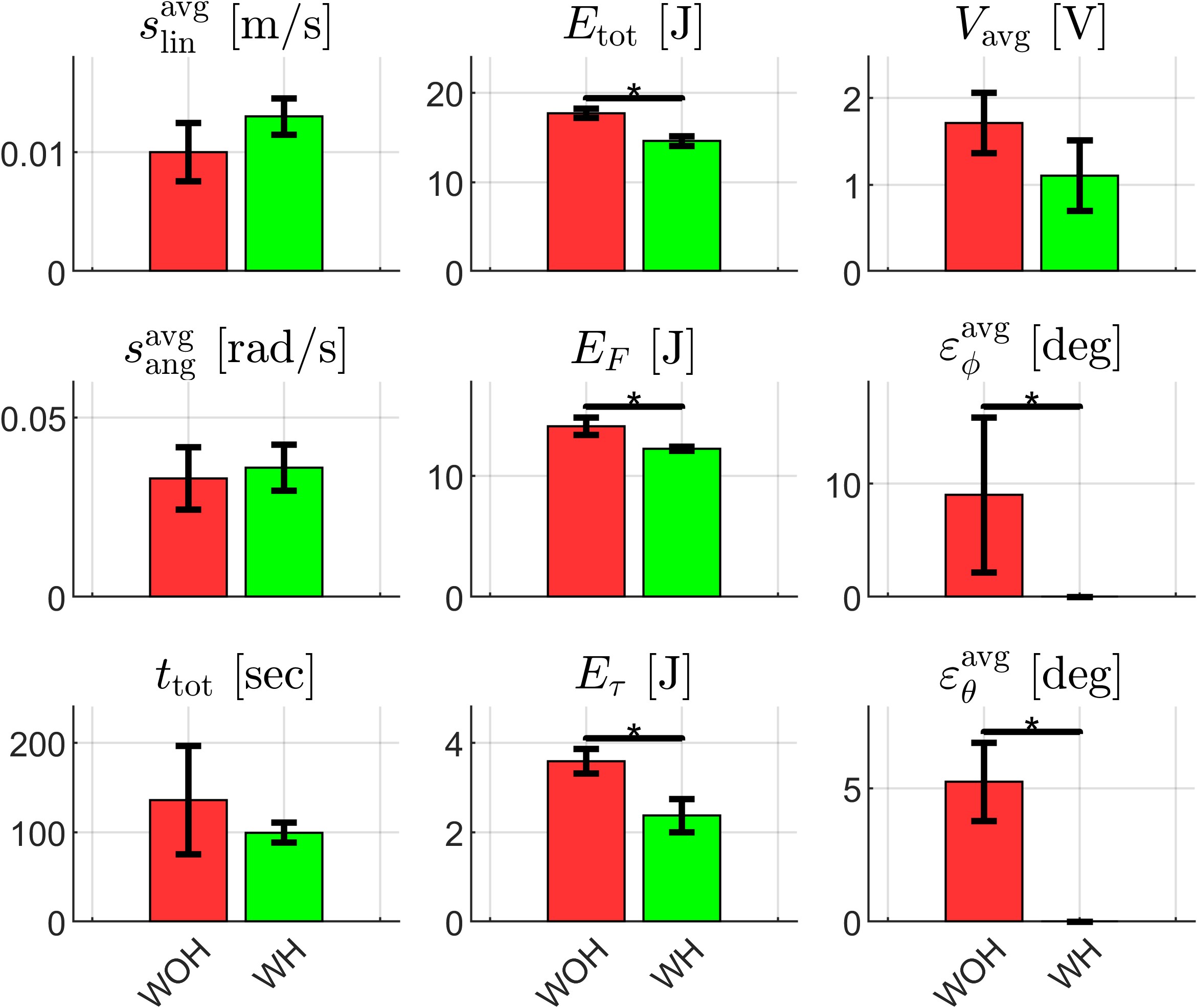}
	\centering
	\caption{Performance metrics for Experiment I, with and without haptic guidance (labeled as ``WH" and ``WOH" respectively). Vertical bars show population means, and vertical lines show 95\% confidence intervals. Horizontal lines with asterisks on top show statistically significant pairwise comparisons with $p=0.05$ as the threshold.}
	\label{fig:exp1quantresults}
\end{figure}

\subsection{Results}

The quantitative metrics used for comparing the experimental conditions (with and without haptic guidance) are the following:

\begin{itemize}
    \item Task completion time, $t_\mathrm{tot}$ [sec]
    \item Average linear speed, $s_\mathrm{lin}^\mathrm{avg} = \frac{1}{t_\mathrm{tot}} \int_0^{t_\mathrm{tot}} ||\mathbf{v}_\mathrm{tool}|| dt$ [m/s]
    \item Average angular speed, $s_\mathrm{ang}^\mathrm{avg} = \frac{1}{t_\mathrm{tot}} \int_0^{t_\mathrm{tot}} ||\mathbf{\omega}_\mathrm{tool}|| dt$ [rad/s]
    \item Average Flexor Carpi muscle activation, $V_\mathrm{avg} = \frac{1}{t_\mathrm{tot}} \int_0^{t_\mathrm{tot}} v(t) dt$ [Volts]
    \item Human effort caused by forces, $E_F$ [J]
    \item Human effort caused by torques, $E_\tau$ [J]
    \item Total human effort, $E_\mathrm{tot}$ [J] = $E_F$ + $E_\tau$
    \item Average alignment error,
    $\varepsilon_\phi^\mathrm{avg}$ for polar, and $\varepsilon_\theta^\mathrm{avg}$ for azimuth angle [deg]
\end{itemize}

Task completion time $t_\mathrm{tot}$ refers to the entire time span of a drilling session where the participant has to use the robot and the AR interface to drill 3 holes on the curved surface. Since it takes 4 seconds per hole (Table \ref{tab:6DexpDetails}) for the robot to align the drill to the desired orientation under the haptic guidance condition, a total of 12 seconds was spent by the robot for the alignment process. Linear (angular) Speed $s_\mathrm{lin}^\mathrm{avg}$ ($s_\mathrm{ang}^\mathrm{avg}$) is simply the mean magnitude of the linear (angular) velocity vector $\mathbf{v}_\mathrm{tool}$ ($\mathbf{\omega}_\mathrm{tool}$) of the drill bit throughout one full session. Average alignment errors $\varepsilon_\phi^\mathrm{avg}$ and $\varepsilon_\theta^\mathrm{avg}$ are calculated at the time of drilling, just before the drilling starts, for every one of the three points. This metric is obviously close to zero when the haptic guidance is enabled since the robot automatically brings the drill to the desired orientation. To calculate the total human effort, $E_\mathrm{tot}$, first, each component of the wrench vector, $\textbf{F}_h = [F_h^x,F_h^y,F_h^z,\tau_h^x, \tau_h^y,\tau_h^z]$, is multiplied with the corresponding component of the velocity vector, $\textbf{v} = [v_x,v_y,v_z,\omega_x, \omega_y,\omega_z]$, and sum of the absolute value of aforementioned element-wise products are computed which gives us the instantaneous human power. Ultimately, this instantaneous power is integrated over time, $E_\mathrm{tot} = \int_t\sum_{i=1}^{6}{\abs{{F}_{h}^i {v}^i} dt}$. Average muscle activation $V_\mathrm{avg}$ is simply the mean activation level of the Flexor Carpi muscle as measured by the sEMG sensors, throughout the trial.

%
%


%

Fig. \ref{fig:exp1quantresults} shows the average task performances of the participants under both experimental conditions based on the aforementioned metrics. To test the null hypothesis, one-way ANOVA (Analysis of Variance) with repeated measures was performed by considering the experimental conditions (with/without haptic guidance) as the main factor. According to the ANOVA results, the effect of the experimental conditions was significant on human effort (the middle column in Fig. \ref{fig:exp1quantresults}) and average alignment errors ($\varepsilon_\phi^\mathrm{avg}$ and $\varepsilon_\theta^\mathrm{avg}$). Specifically, both linear and angular components of the human effort were significantly lower under haptic guidance. The alignment errors were between 3$^\circ$ and 16$^\circ$ without haptic guidance. According to Fig. \ref{fig:exp1quantresults}, average speeds, $s_\mathrm{lin}^\mathrm{avg}$ and $s_\mathrm{ang}^\mathrm{avg}$, are generally higher under haptic guidance (though the difference between the conditions is not statistically significant), leading to a lower task completion time $t_\mathrm{tot}$. The same applies to average muscle activation $V_\mathrm{avg}$, whose population mean is lower under haptic guidance, but the difference between the conditions is not statistically significant. A large amount of variance can be observed in task completion time $t_\mathrm{tot}$ in Fig. \ref{fig:exp1quantresults} when not using haptic guidance since the whole alignment is done manually by the participant, and the time it takes to do the alignment can vary considerably between the trials of the same participant and also between the participants.

\section{EXPERIMENT II}
\label{sec:exp2}

In this experiment, the objective was to examine the acceptability of the proposed pHRI system and the drilling scenario, if deployed in real manufacturing settings and used by real industrial workers. For this reason, three industrial workers with extensive experience in small-batch manufacturing tasks including drilling were invited to our lab for experimentation. They were asked to open holes on the curved surface under haptic guidance condition only. The target locations and drilling angles were the same as the ones used in Experiment I (Table \ref{tab:6DexpDetails}). Following the experiment, the participants were asked to fill out a detailed questionnaire designed to acquire their subjective opinions about the developed system and its applicability in real manufacturing settings.

\subsection{Protocol}

The protocol followed in this experiment is identical to the one mentioned in \hyperref[sec:exp1]{Section 3} for the haptic guidance condition. 
After the training, every participant performed one trial (3 holes) of the experiment under haptic guidance only.

\subsection{Questionnaire}

Methods suggested by \cite{Basdogan2000,Kucukyilmaz2013} were utilized for designing a subjective questionnaire\footnote{Full text of the questionnaire can be found at \url{https://www.yusuf-aydin.com/iros-2022-robot-assisted-drilling/\#questionnaire}} that utilizes a 7-point Likert scale for collecting the personal opinions of the participants about the developed pHRI system. A series of statements were put forth in the questionnaire, with which the participants can specify to what extent they agree. The 7-point Likert scale varied from a scale of 1 (strongly disagree) to 7 (strongly agree). The statements were arranged in a fixed but shuffled order, and asked twice with different wordings, meaning every statement also has a paraphrased opposite pair in the questionnaire. 

In the first part of the questionnaire, demographic data were collected from the participants about their gender, age, and years of work experience in the manufacturing industry. This section also included a question asking whether the participant believes drilling holes on a curved surface using a manual drill is difficult, and if so, why. 

In the second section of the questionnaire, first, four general questions were asked about how much prior experience the participant had with mechanical tools (such as screwdrivers, chain saws, etc.), robots, and AR/VR interfaces in general. Then, 11 questions about our pHRI system and 3 questions about our AR interface were asked in particular, along with their opposite pairs. Therefore, this section of the questionnaire included a total of $2\times(11+3) + 4 = 32$ questions. The questions in this section were related to the following topics:

\begin{itemize}
    \item Personal prior experience with: non-motorized devices (hammer, screwdriver, etc.), motorized devices (drill, chain saw, polisher, etc.), robotic arms, augmented/virtual reality interfaces
    \item Using the robot in the experiments: Overall task success, ease of use, being in control, task performance, trust, feeling natural, enjoyment, safety, fear, exhaustion, willingness to use again
    \item Using the AR interface in the experiments: Ease of use, enjoyment, and utility (helpfulness)
\end{itemize}

\begin{figure}[h]
	\includegraphics[width=\columnwidth]{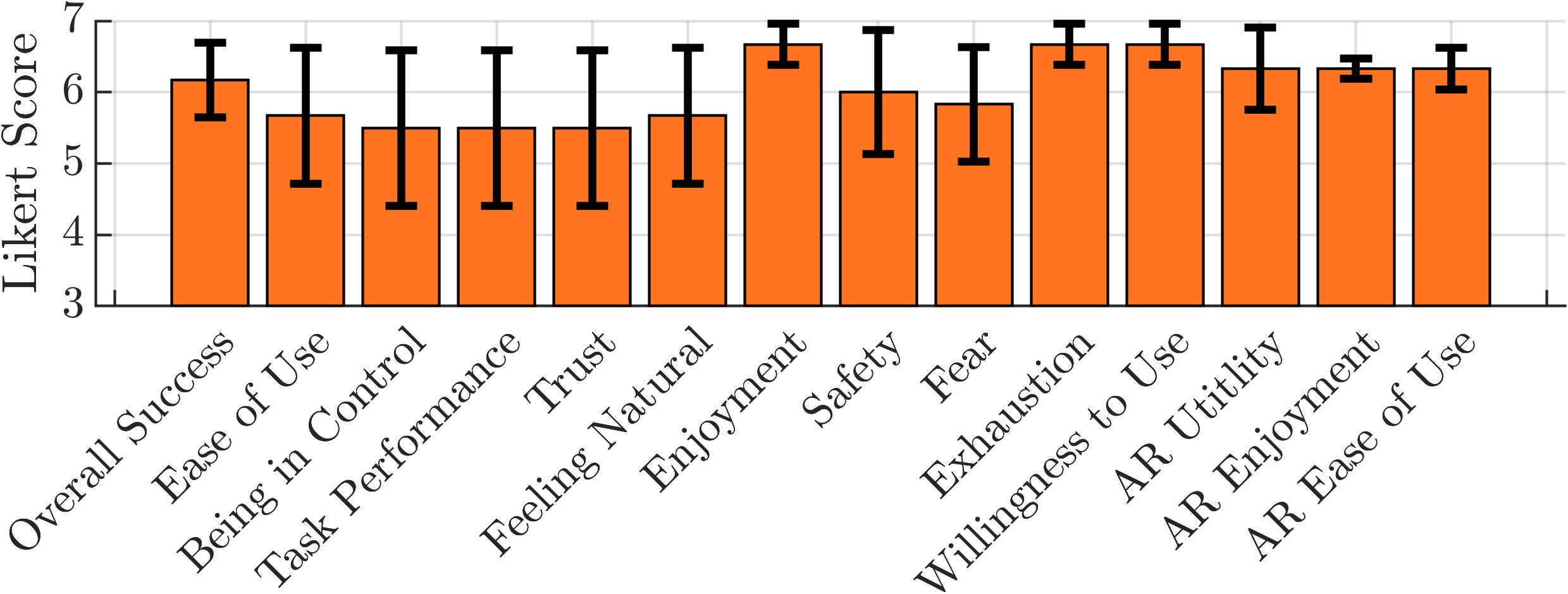}
	\centering
	\caption{Subjective questionnaire results in Experiment II. Bars indicate means and vertical lines indicate standard deviations.}
	\label{fig:questionnaire}
\end{figure}

\subsection{Results}

\subsubsection{Personal Background}

The average age of the workers was 31.67 $\pm$ 8.62 years. They had 15.67 $\pm$ 9.02 years of experience in the manufacturing industry. None of the participants stated that they encountered any problem during the experiments. 2 out of the 3 participants stated that drilling holes on curved surfaces with custom drilling angles \textcolor{black}{is} hard because it is difficult to manually align a drill to a defined orientation by hand. All 3 participants stated that they had extensive experience with non-motorized and motorized mechanical devices (Strongly Agree; 7.0/7.0). The average Likert score for experience with robotic arms was 6.7/7.0, indicating that the participants had worked with robotic arms before. The mean score for prior experience with AR interfaces was 3.7/7.0. The participants later indicated that they became familiar with AR/VR interfaces as they have attended technical fairs and exhibits.

\subsubsection{Subjective Questionnaire}

Fig. \ref{fig:questionnaire} shows the mean responses of the 3 participants to the questions. Despite low scores occasionally, all questions were scored higher than 5.0/7.0 on average by the participants, suggesting that they generally had a positive \textcolor{black}{opinion} about our pHRI system.

According to the results shown in Fig. \ref{fig:questionnaire}, the participants on average found the robot easy to use, safe, convenient, and natural. They were not afraid of using the robot, and were willing to use it again. The questions about the AR interface received scores on average higher than 6.0/7.0, suggesting that the participants found the AR goggle to be \textcolor{black}{user-friendly}, and helpful in carrying out the task.

%


\section{DISCUSSION and CONCLUSION}
\label{sec:conclusion}
In this study, a pHRI system was developed to perform small-batch manufacturing tasks 
efficiently using haptic guidance, an AR interface, and an adaptive admittance controller.
As a case study, collaborative drilling of holes on a curved surface was selected. The proposed system provides haptic guidance to the user by constraining the user's movement to the drilling axis and visual guidance through the AR interface for the steps of the drilling task and rough alignment of the drill bit to the desired angle. Furthermore, the damping parameter of the admittance controller that regulates the interaction between the robot and the human was adapted based on the instantaneous position of drill tip with respect to the location of drill target. During the free motion, in which the user brings the drill \textcolor{black}{to the 5-cm vicinity of} target point, low admittance damping was utilized so that the robot showed minimal resistance to the human. On the other hand, high admittance damping was used during the drilling phase for stable and safe operation.

The benefits of haptic guidance were investigated in Experiment I by comparing the task performance of the participants with and without haptic guidance. In order to compensate for the lack of haptic guidance under no haptic guidance condition, more visual guidance was provided to the participants through the AR interface. Since the participants manually adjusted the drill bit by themselves under no haptic guidance condition, the angular deviations of drill bit from the desired drill orientation were displayed through the AR interface and guiding arrows were used to help the participant make corrections for those deviations. 

\begin{figure}[h]
    \vspace{10pt}
	\includegraphics[width=\columnwidth]{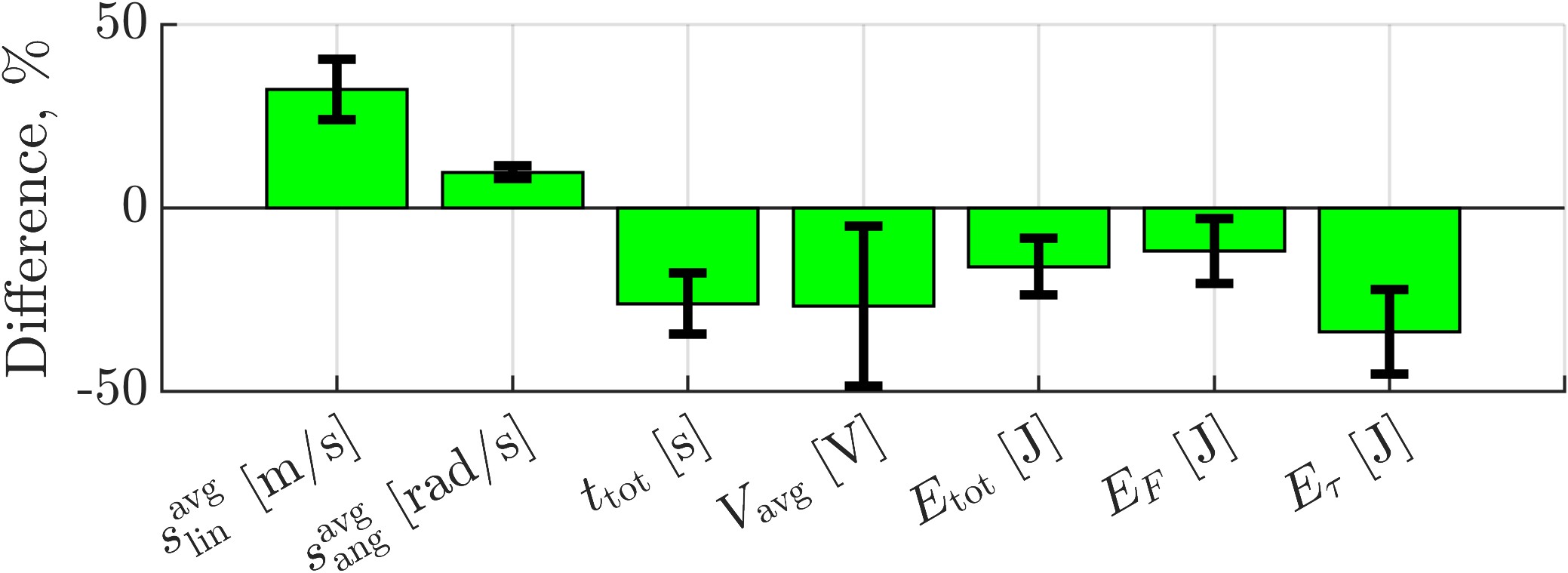}
	\centering
	\caption{Percent relative difference in the performance metrics of the participants under the experimental conditions of with and without haptic guidance in Experiment I. The negative (positive) values in the plot indicate decrease (increase) in the metrics of haptic guidance condition with respect to the no haptic guidance condition. Vertical bars show average differences among the three subjects, and vertical lines indicate their standard deviations.}
	\label{fig:percentdiff}
	  \vspace{-1.5em}
\end{figure}

The results of Experiment I showed that the task performance under haptic guidance was superior to that of the no haptic guidance condition. Fig. \ref{fig:percentdiff} presents the relative differences in percentage between the performance metrics of the two conditions. According to this figure, the haptic guidance condition led to 26\% shorter task completion time, and 16\% less human effort. Furthermore, 27\% less muscle activation was observed in the participants under the haptic guidance condition, suggesting that haptic guidance could potentially improve the ergonomics in pHRI tasks by decreasing the chances of muscular fatigue in the users.

The focus of Experiment II was to test the potential use of the proposed system in real manufacturing environments based on the subjective opinions of industrial workers. For this reason, Experiment II was conducted with 3 industrial workers who had extensive experience in performing small-batch tasks such as drilling. The results of Experiment II showed that the workers found the system easy to use and helpful in carrying out the drilling task. Also, the results showed that they enjoyed the system, felt safe and successful, and were willing to use it again.  


For future work, a depth camera can be attached to the robot for extracting surface information locally only rather than for the whole surface, as was done in our study. Moreover, the workpiece could be tracked in real time, thereby eliminating the need to rigidly fix it in place. Consequently, a significant amount of time can be saved when working on a part with unknown geometry \textcolor{black}{held in human hand}.





\section*{ACKNOWLEDGMENT}

A.M., P.P.N., and B.G. acknowledge the research fellowship provided by the KUIS AI-Center. The authors would like to thank the project's partner company, As-Metal Inc., for providing the curved workpieces used in both experiments and the industrial workers participated in the second experiment.

\bibliography{references}

\end{document}